\documentclass[10pt, conference, compsocconf]{IEEEtran}
\ifCLASSINFOpdf
\else
\fi
\usepackage{amsmath}
\usepackage{url}
\usepackage{multirow}
\usepackage[font={normalsize}]{caption}
\usepackage{graphicx}

\usepackage{cite}

\usepackage{float}
\floatstyle{plaintop}
\restylefloat{table}

\newtheorem{definition}{Definition}
\newtheorem{example}{Example}

\begin{document}
%
\title{CER: Complementary Entity Recognition via Knowledge Expansion\\on Large Unlabeled Product Reviews}


\author{
    \IEEEauthorblockN{Hu Xu\IEEEauthorrefmark{1}, Sihong Xie\IEEEauthorrefmark{2}, Lei Shu\IEEEauthorrefmark{1}, Philip S. Yu\IEEEauthorrefmark{1}\IEEEauthorrefmark{3}}
    \IEEEauthorblockA{\IEEEauthorrefmark{1}Department of Computer Science, University of Illinois at Chicago, Chicago, IL, USA}
    \IEEEauthorblockA{\IEEEauthorrefmark{2}Department of Computer Science and Engineering, Lehigh University, Bethlehem, PA, USA}
    \IEEEauthorblockA{\IEEEauthorrefmark{3}Institute for Data Science, Tsinghua University, Beijing, China}
	\IEEEauthorblockA{hxu48@uic.edu, sxie@cse.lehigh.edu, lshu3@uic.edu, psyu@uic.edu}
}


%


\maketitle

\begin{abstract}
Product reviews contain a lot of useful information about product features and customer opinions. One important product feature is the \emph{complementary entity} (products) that may potentially work together with the reviewed product. Knowing complementary entities of the reviewed product is very important because customers want to buy compatible products and avoid incompatible ones. In this paper, we address the problem of Complementary Entity Recognition (CER). Since no existing method can solve this problem, we first propose a novel unsupervised method to utilize syntactic dependency paths to recognize complementary entities. Then we expand category-level domain knowledge about complementary entities using only a few general seed verbs on a large amount of unlabeled reviews. The domain knowledge helps the unsupervised method to adapt to different products and greatly improves the precision of the CER task. The advantage of the proposed method is that it does not require any labeled data for training. We conducted experiments on 7 popular products with about 1200 reviews in total to demonstrate that the proposed approach is effective.
\end{abstract}


%
\IEEEpeerreviewmaketitle

\section{Introduction}
\label{sec:intro}
\begin{figure}[tbp] 
   \centering
   \includegraphics[width=3.2in]{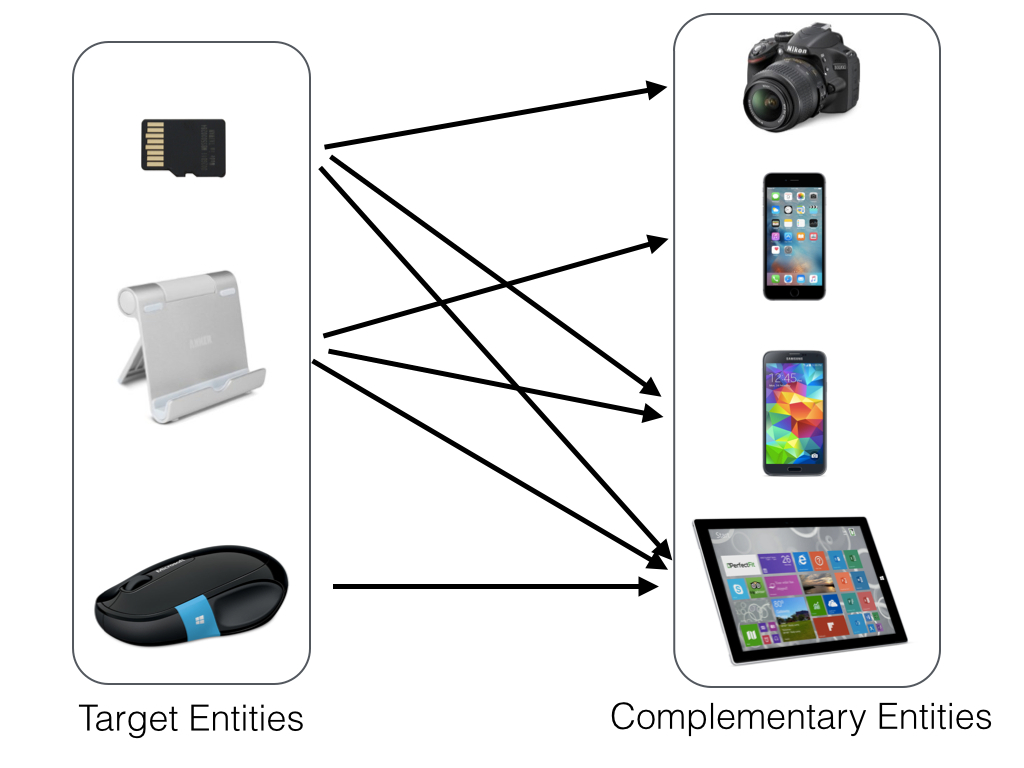}
   \caption{Several target entities (reviewed products), their complementary entities and complementary relations mentioned in reviews.}
   \label{fig:sc}
\end{figure}

E-commerce websites (e.g., Amazon.com) contain a huge amount of products reviews and most existing works of sentiment analysis \cite{pang2002thumbs} (or opinion mining) on reviews focus on extracting opinion targets (aspects or features) of the reviewed product and the associated opinions \cite{hu2004mining,popescu2007extracting,liu2015sentiment} (e.g., extract ``battery'' and \textit{pos} from ``It has a good battery''). Besides features about the reviewed product itself (e.g., ``battery'' or ``screen''), one important feature is whether the reviewed product is compatible/incompatible with another product. We call the reviewed product \emph{target entity} and the other product \emph{complementary entity}. A pair of a target entity and its complementary entity forms a \emph{complementary relation}. They may work together to fulfill some shared functionalities. So, they are usually co-purchased. For example, in Figure \ref{fig:sc}, we assume there are some reviews of several accessories (on the left) talking about compatibility issues. We consider these accessories as the target entities and they have some complementary entities (on the right side) mentioned in reviews. The target entities are one \textit{micro SD card}, one \textit{tablet stand} and one \textit{mouse}; the complementary entities are one \textit{Nikon DSLR}, one \textit{iPhone}, one \textit{Samsung Galaxy S6} and one \textit{MS Surface Pro}. An arrow pointing from a target entity to a complementary entity indicates that they have a complementary relation and shall work together. For example, the \textit{micro SD card} can help the \textit{Samsung Galaxy S6} to expand its memory capacity. Knowing these complementary entities is important because compatible products are preferred over incompatible ones. Thus, recognizing complementary entities is an important task in text mining.

\textbf{Problem Statement}: In this paper, we study the problem of Complementary Entity Recognition (CER) from reviews (e.g., extracting ``Samsung Galaxy S6'' from ``It works with my Samsung Galaxy S6'' ). We observe that compatibility issues are more frequently discussed in reviews of electronics accessories, so we choose reviews of accessories for experiments. To the best of our knowledge, accessory reviews are not well studied before.

Predicting complementary entities is pioneered by McAuley et al. \cite{McAPanLes15} as a link prediction problem in social network. Their method mostly predicts category-level compatible products based on the learned representations of the products. However, we observe that reviews contain many complementary entities based on firsthand user experiences, which provide practical fine-grained complementary entities. We detail the discussions of their method in Section \ref{sec:rw}.

The proposed problem has a few challenges and also provides more research opportunities:
\begin{itemize}
	\item To the best of our knowledge, the linguistic patterns of complementary relations are not studied in computer science. There is no largely annotated dataset for supervised methods. We propose an unsupervised method, which does not require any labeled data to solve this problem (we only annotate a small amount of data for evaluation purposes).
	\item Similar to the aspect (feature) extraction problem in reviews \cite{liu2015sentiment}, CER is also a domain-specific problem. We leverage domain knowledge to help the unsupervised method to adapt to different products. This novel product domain knowledge is expanded using a few seed words on a large amount of unlabeled reviews under the same category as the target entity. The idea of using reviews under the same category as the target entity is that the number of reviews for one target entity is small. We observe that products (target entities) under the same category share similar complementary entities (i.e., two different \textit{micro SD card}s may share complementary entities like \textit{phone} or \textit{tablet}). So the domain knowledge expanded on reviews from the same category is larger than that on reviews from a single target entity. Therefore, there is almost no labor-intensive effort to get domain knowledge. Our domain knowledge contains candidate complementary entities and domain-specific verbs.
	\item Although the problem may be closely related to the well-known Named Entity Recognition (NER) problem on surface \cite{nadeau2007survey}, recognizing a complementary entity requires more contexts. For example, given a review for a \textit{micro SD card}, we should not treat ``Samsung Galaxy S6'' in ``Samsung Galaxy S6 is great'' as a complementary entity. However, we should consider the same entity in ``It works with my Samsung Galaxy S6'' as a complementary entity. The domain knowledge contains domain-specific verbs, which greatly help to detect the contexts of complementary entities.
	\item We further notice that some linguistic patterns of complementary relations are similar to other extraction patterns (e.g., patterns for aspect extraction). Candidate complementary entities in the domain knowledge can help to filter out non-complementary entities extracted by similar patterns.
\end{itemize}

The main contributions of this paper can be summarized as the following: we propose a novel problem called  Complementary Entity Recognition (CER). Then we propose a novel unsupervised method utilizing dependency paths to identify complementary relations and extract entities simultaneously. We further leverage domain knowledge to improve the precision of extraction. The domain knowledge is expanded on a large amount of unlabeled reviews from only a few seed words (general complementary verbs) via a novel set of dependency paths. The expanded domain knowledge can greatly improve the precision of the unsupervised method. We conduct thorough experiments and provide case studies to demonstrate that the proposed method is effective.


\section{Related Works}
\label{sec:rw}
The proposed problem is closely related to product recommender systems that are able to separate substitues and complements  \cite{McAPanLes15,zheng2009substitutes}. Zheng et al. \cite{zheng2009substitutes} first propose to incorporate the concepts of substitutes and complements into recommendation systems by analyzing navigation logs. More specifically, predicting complementary relations is pioneered by McAuley et al. \cite{McAPanLes15}. They utilize topic models and customer purchase information (e.g., the products in the ``items also viewed'' section and the ``items also bought'' section of a product page) to predict category-level substitutes and complements. However, we observe that purchase information generated by the unknown algorithm from Amazon.com tends to be noisy and inaccurate for complementary entities since co-purchased products may not be complementary to each other. We demonstrate that their predictions are non-complementary entities for the products that we use for experiments in Section \ref{sec:exp}. Also, category-level predictions are not good enough for specific pairs of products (i.e., \textit{DSLR lens} and \textit{webcam} are not complements). Furthermore, their predictions do not provide information about incompatible entities, which are valuable buying warnings for customers. Thus, fine-grained extraction of complementary entities from reviews that express firsthand user experience is important. To the best of our knowledge, the linguistic patterns of complementary relations are not studied in computer science.

The proposed problem is closely related to aspect extraction \cite{hu2004mining,popescu2007extracting,qiu2011opinion,liu2015sentiment}, which is to extract product features from reviews. More specifically, extracting comparable products (i.e, one type of substitutes, or products that can replace each other) from reviews is studied by Jindal and Liu \cite{jindal2006mining}. Recently, dependency paths \cite{kubler2009dependency} are used for aspect extraction \cite{qiu2011opinion,liu2015automated}. Shu et al. \cite{shu2016lifelong} use unsupervised graph labeling method to identify entities from opinion targets. However, since aspects are mostly context independent and the same aspect may appear multiple times, aspect extraction in general does not need to extract each occurrence of an aspect (as long as the same aspect can be extracted at least once). In contrast, the CER problem is context dependent and many complementary entities are infrequent (i.e., \textit{Samsung Galaxy S6} is infrequent than the aspect \textit{price}). We use dependency paths to accurately identify each occurrence of complementary entities. Since extracting each complementary entity can be inaccurate, we further utilize domain knowledge to improve the precision. 

CER is closely related to Named Entity Recognition (NER)\cite{nadeau2007survey} and relation extraction \cite{bach2007review}. NER methods utilize annotated data to train a sequential tagger \cite{rabiner1986introduction,mccallum2000maximum,lafferty2001conditional}. However, our task is totally different from NER since we care about the context of a complementary entity and many complementary entities are not named entities (e.g., \textit{phone}). CER is also different from relation extraction \cite{bach2007review,culotta2004dependency,mintz2009distant,bunescu2005shortest}, which assumes that two entities are identified in advance. In reviews, the target entity is unfortunately missing in many cases (i.e., ``Works with my phone''). The proposed method only cares about the relation context of a complementary entity rather than a full relation.

\section{Preliminaries}
\label{sec:prelim}
In this section, we first formally define our problem. Then we introduce basic ideas of the proposed method. Lastly, we describe dependency paths used in later sections.

\subsection{Problem Formalization}
Our problem is to recognize entities that functionally complement to the reviewed product. There are several definitions involved in this problem.

\begin{definition} [Target Entity] \label{defn:te}
We define \emph{target entity} $e_T$ as the reviewed product.
\end{definition}

We do not extract target entities from reviews but assume that the target entity can be retrieved from the meta data (product title) of reviews. This is because many mentions of the target entity are co-referenced or implicitly assumed in reviews. For example, if the reviewed product is a \textit{tablet stand}, ``It works with my Samsung Galaxy S6'' uses ``It'' to refer to the target entity \textit{tablet stand}; ``Works well with Samsung Galaxy S6'' completely omits the target entity.

\begin{definition} [Complementary Entity] \label{defn:ce}
Given a set of reviews $R_T$ of a target entity $e_T$, a \emph{complementary entity} $e_C$ is an entity mentioned in reviews that are functionally complementary to the target entity $e_T$. A target entity has a set of complementary entities: $e_C \in E_C$.  
\end{definition}

A complementary entity can either be a single noun (e.g., \textit{iPhone}) or a noun phrase (e.g., \textit{Samsung Galaxy S6}). There are two types of complementary entities: a \emph{named entity} or a \emph{general entity}. A named entity is usually a specific product name containing a brand name and a model name (e.g., \textit{Samsung Galaxy S6} or \textit{Apple iPhone}). A general entity (e.g., \textit{phone} or \textit{tablet}) represents a set of named entities. General entities are informative. For example, in a review of a \textit{tablet stand}, ``phone'' in ``It also works with my phone'' is a good assurance for phone owners who want to use this \textit{tablet stand} as a \textit{phone stand}.

\begin{definition} [Complementary Relation] \label{defn:cr} 
Each complementary entity $e_C \in E_C$ forms a \emph{complementary relation} $(e_T, e_C)$ with the target entity $e_T$. 
\end{definition}

\begin{definition} [\textbf{Complementary Entity Recognition}] \label{defn:cee}
Given a set of reviews $R_T$ for a target entity $e_T$, the problem of Complementary Entity Recognition (CER) is to identify a set of complementary entities $E_C$, where each $e_C \in E_C$ has a complementary relation $(e_T, e_C)$ with the target entity $e_T$.
\end{definition}

We do not extract an entity without a complementary context (e.g., ``Samsung Galaxy S6'' in ``Samsung Galaxy S6 is great'', even though \textit{Samsung Galaxy S6} may be a complementary entity).

\begin{definition} [Domain] \label{defn:domain}
We assume that every target entity $e_T$ belongs to a pre-defined \emph{domain} (or \emph{category}) $\textit{Dom}(e_T)=d \in D$. A \emph{review corpora} $R^{\textit{Dom}(e_T)}$ is all reviews under the same category as the target entity $e_T$.
\end{definition}

\begin{definition} [Domain Knowledge] \label{defn:domainknowledge} 
Each domain $d$ has its own \emph{domain knowledge}. We consider two types of domain knowledge: \emph{candidate complementary entity} $e_C^d \in E_C^d$ and \emph{domain-specific verb} $v^d \in V^d$. All target entities $e_T$ under the same domain share the same domain knowledge.
\end{definition}

\subsection{Basic Ideas}
The basic idea of the proposed method is to use dependency paths to identify complementary entities. Due to different linguistic patterns, these dependency paths may have different performance on extraction. Some dependency paths may have high precision but low recall and vice versa. To ensure the quality of extraction, high precision dependency paths are preferred. The idea of using domain knowledge is that high precision dependency paths can expand high quality (precision) domain knowledge on a large amount of unlabeled reviews, which in turn helps low precision but high recall dependency paths to improve their precisions. In the end, the domain knowledge serves as a filter to remove noises in low precision paths. This framework can potentially be generalized to any extraction task when a large amount of unlabeled data is accessible. We describe the proposed method in the following two parts:\\
\textbf{Basic Entity Recognition}: We analyze the linguistic patterns and leverage multiple dependency paths to recognize complementary entities. The major goal of the basic entity recognition is to get high recall because each complementary entity can be infrequent and we care about each mention of a complementary entity. Due to similarity with other noisy patterns, these paths tend to have a low precision.\\
\textbf{Recognition via Domain Knowledge Expansion}: We expand the domain knowledge on a large amount of unlabeled reviews using a set of high precision dependency paths to compensate for the low precision (noisy) dependency paths. First, we extract candidate complementary entities for each domain using only verbs \textit{fit} and \textit{work}. Then we use the extracted candidate complementary entities to induce domain-specific verbs (e.g., \textit{insert} for \textit{micro SD card}, or \textit{hold} for \textit{tablet stand}). Finally, we integrate these two types of domain knowledge into the dependency paths of basic entity recognition to improve the precision.

\subsection{Dependency Paths}
In this subsection, we briefly review the concepts used by dependency paths. We further describe how to match a dependency path with a sentence.
 
\begin{definition} [Dependency Relation] \label{defn:dr} 
A \textit{dependency relation} is a typed relation between two words in a sentence with the following format of \emph{attributes}: 
$$\textit{type(gov, govidx, govpos, dep, depidx, deppos)}, $$
where \textit{type} is the type of a dependency relation, \textit{gov} is the \emph{governor word}, \textit{govidx} is the index (position) of the \textit{gov} word in the sentence, \textit{govpos} is the POS (Part-Of-Speech) tag of the \textit{gov} word, \textit{dep} is the \emph{dependent word}, \textit{depidx} is the index of the \textit{dep} word in the sentence and \textit{deppos} is the POS tag of the \textit{dep} word. The \emph{direction} of a dependency relation is from the \textit{gov} word to the \textit{dep} word.
\end{definition}

\begin{figure}[htbp] 
   \centering
   \includegraphics[width=3in]{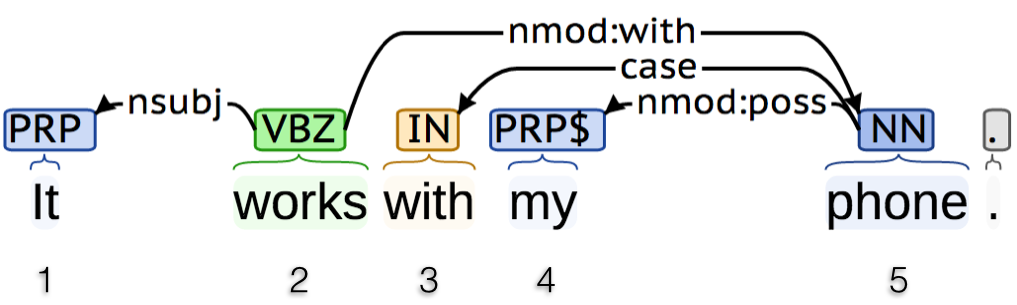} 
   \caption{Visualization of dependency relations of ``It works with my phone'': numbers indicates indices.}
   \label{fig:dt}
\end{figure}

A sentence can be parsed into a set of dependency relations through dependency parsing\footnote{We utilize Stanford CoreNLP as the tool for dependency parsing.} \cite{kubler2009dependency,de2008stanford}. For example, ``It works with my phone'' can be parsed into a set of dependency relations in Table \ref{table:dr}, which is further illustrated in Figure \ref{fig:dt}.

\begin{table*}[ht]
\centering
\scalebox{0.95}{
\begin{tabular}{ c | c | c | l }
\hline
ID & Dependency Relation & Syntactic Dependency Relation Type & Explanation \\ 
\hline
1 & \textit{nsubj(works, 2, VBZ, It, 1, PRP)} & \textit{nsubj}: nominal subject & \begin{tabular}[t]{@{}l@{}} Relate the 1st word ``It'' \\to the 2nd word ``works'' \end{tabular} \\
\hline
2 & \textit{root(ROOT, 0, None, works, 2, VBZ)} & \textit{root}: root relation & \begin{tabular}[t]{@{}l@{}} Relate the 2nd word ``works''\\to the virtual word ROOT\end{tabular} \\
\hline
3 & \textit{case(phone, 5, NN, with, 3, IN)} & \textit{case}: case-marking & \begin{tabular}[t]{@{}l@{}} Relate the 3rd word ``with'' \\to the 5th word ``phone'' \end{tabular} \\
\hline
4 & \textit{nmod:poss(phone, 5, NN, my, 4, PRP\$)} & \textit{nmod:poss}: possessive nominal modifier & \begin{tabular}[t]{@{}l@{}} Relate the 4th word ``my'' \\to the 5th word ``phone''\end{tabular} \\
\hline
5 & \textit{nmod:with(works, 2, VBZ, phone, 5, NN)} & \textit{nmod:with}: nominal modifier via with &  \begin{tabular}[t]{@{}l@{}} Relate the 5th word ``phone'' \\to the 2nd word ``works'' \end{tabular} \\
\hline 
\end{tabular}
}
\caption{Dependency relations parsed from ``It works with my phone.''}
\label{table:dr}
\end{table*}

\begin{definition} [Dependency Segment] \label{defn:seg} 
\emph{A dependency segment} is an abstract form of a dependency relation. A dependency segment has the following format of attributes, which is similar to a dependency relation: 
$$(\textit{src, srcpos})\xrightarrow[]{\textit{pathtype}}(\textit{dst, dstpos}), $$
where \textit{src} is the \textit{source word}, \textit{srcpos} is the POS tag of the source word, \textit{dst} is the \textit{destination word}, \textit{dstpos} is the POS tag of the destination word and \textit{pathtype} is the \textit{dependency type} of the segment. Similarly, the \emph{direction} of an segment is from the \textit{src} word to the \textit{dst} word.
\end{definition}

\begin{definition} [Dependency Segment Matching] \label{defn:match} 
A dependency segment can have a \emph{dependency segment matching} with a dependency relation. To have such a match, we must ensure that attributes \textit{src}, \textit{srcpos}, \textit{dst}, \textit{dstpos} and \textit{pathtype} in an segment match attributes \textit{gov}, \textit{govpos}, \textit{dep}, \textit{deppos} and \textit{type} in a dependency relation respectively. So the direction of a dependency segment also matches the direction of a dependency relation.
\end{definition}

To allow a matching to cover more specific dependency relations, we further define a set of rules when matching the attributes, which are summarized in Table \ref{table:match}. Please note that we finally want to extract the complementary entity covered by tag \textit{CETT}. Other kinds of attributes are defined to make the dependency paths more compact.

\begin{table}
\scalebox{0.88}{
\begin{tabular}{ l | c | c | c }
\hline
Path Attr. & Value & Rel. Attr. & Value\\
\hline
\textit{src/dst} & [\textit{lem. word}] & \textit{gov/dep} & [\textit{specific form}]\\
\hline
\textit{src/dst} & \textit{* / CETT} & \textit{gov/dep} & [\textit{any word}]\\
\hline
\textit{srcpos/dstpos} & \textit{N} & \textit{gov/dep} & 
\begin{tabular}[t]{@{}l@{}} 
\textit{NN} \textit{NNP} \textit{NNPS} \textit{NP}
\end{tabular}\\
\hline
\textit{srcpos/dstpos} & \textit{V} & \textit{gov/dep} & 
\begin{tabular}[t]{@{}l@{}}
\textit{VB} \textit{VBD} \textit{VBG}\\
\textit{VBN} \textit{VBP} \textit{VBZ}
\end{tabular}\\
\hline
\textit{srcpos/dstpos} & \textit{J} & \textit{gov/dep} & \textit{JJ} \textit{JJR} \textit{JJS}\\
\hline
\textit{pathtype} & \textit{nmod:cmprel} & \textit{type} & 
\begin{tabular}[t]{@{}l@{}} 
\textit{nmod:with} \textit{nmod:for} \\
\textit{nmod:in} \textit{nmod:on} \\
\textit{nmod:to} \textit{nmod:inside} \\
\textit{nmod:into}
\end{tabular}\\
\hline 
\end{tabular}
}
\caption{Rules of matching attributes of dependency segments and dependency relations (all unspecified attributes must have an exact match): [\textit{lem. word}] means lemmatized word, which matches multiple specific forms of the same word (e.g., ``work'' matches ``works'' and ``working''); \textit{CETT} indicates the complementary entity we want to extract.}
\label{table:match}
\end{table}

\begin{example}
The segment:  
\begin{equation}
\label{eq:l11}
(\textit{``work'', V})\xrightarrow[]{\textit{nmod:cmprel}}(\textit{CETT, N})
\end{equation}
can match the dependency relation 5 in Table \ref{table:dr}. This is because source word \textit{``work''} is the lemmatized governor word \textit{``works''}; \textit{V} covers \textit{VBZ}; \textit{N} covers \textit{NN}; and \textit{nmod:cmprel} covers dependency type \textit{nmod:with}. Since the tag \textit{CETT} as the destination word in the segment covers the dependent word ``phone'' in dependency relation 5, this segment indicates ``phone'' is a possible complementary entity.
\end{example}

\begin{definition} [Dependency Path] \label{defn:path}
A \textit{dependency path} is a finite sequence of dependency segments connected by a sequence of \textit{src/dst} attributes.
\end{definition}

Given different directions of 2 adjacent dependency segments, there are 4 possible types of a connection: $\rightarrow \rightarrow$, $\rightarrow \leftarrow$, $\leftarrow \rightarrow$ and $\leftarrow \leftarrow$.

\begin{definition} [Dependency Path Matching] \label{defn:match2} 
A procedure of \emph{dependency path matching} is specified as the following: when matching a dependency path with a sentence, we first check whether there are at least one dependency relations for each segment. If so, we further check whether the two directions of dependency segments for each connection match the directions of two corresponding dependency relations and whether the connected governor/dependent words from two matched dependency relations have the same index (they are the same word in the original sentence).
\end{definition}

Finally, after we have a successful dependency path matching, we extract the \textit{gov/dep} in dependency relations labeled as \textit{CETT} by the dependency path.

\begin{example}
The following path
\begin{equation}
\label{eq:l2}
\begin{split}
(\textit{*, V})\xrightarrow[]{\textit{nmod:with}}(\textit{CETT, N})\xrightarrow[]{\textit{nmod:poss}}(\textit{``my'', PRP\$})
\end{split}
\end{equation}
can match the sentence ``It works with my phone'' since the two segments match dependency relation 5 and 4 respectively. Here wildcard \textit{*} matches word ``works''. Further the dependent word ``phone'' of the dependency relation 5 have the same index (the 5th word described in Table \ref{table:dr}) as the governor word of the dependency relation 4. 
\end{example}

\section{Basic Entity Recognition}
\label{sec:r}

\subsection{Syntactic Patterns of Complementary Relation}
There are many ways to mention complementary relations in reviews. Complementary relations are usually expressed with or without a preposition. In the first case, the preposition is used to bring out the complementary entity and is usually associated with a verb, a noun, an adjective or a determiner; in the second case without a preposition, reviewers only use transitive verbs to bring out the complementary entities. The verbs used in both cases can either be general verbs such as ``fit'' or ``work'', or domain-specific verbs such as ``insert'' for \textit{micro SD card} or ``hold'' for \textit{tablet stand}. Complementary relations can also be expressed through nouns, adjectives or determiners. We discuss the syntactic patterns of complementary relations as the following:\\
\textbf{Verb+Prep}: The majority of complementary relations are expressed through a verb followed by a preposition. For example, ``It works with my phone'' falls into this pattern, where the verb ``works'' and the preposition ``with'' work together to relate the pronoun ``It'' to ``phone''. The target entity can appear in this pattern either as the subject or as the object of the verb. In the previous example, subject ``It'' indicates the target entity. In ``I insert the card into my phone'', ``the card'' is the object of the verb ``insert''. The target entity can also be implicitly assumed as in ``Works with my phone.''\\
\textbf{Noun+Prep}: Complementary relation can be expressed through nouns. Those nouns typically have opinions. For example, ``No problem'' in ``No problem with my phone'' has a positive opinion on ``phone''.\\
\textbf{Adjective+Prep}: Complementary relation can also be expressed through adjectives with prepositions. For example, the adjective ``useful'' together with the preposition ``for'' in ``It is useful for my phone'' expresses a positive opinion on a complementary relation.\\
\textbf{Determiner+Prep}: Determiner ``this'' in ``I use this for my phone'' refers to the target entity. It is associated with the preposition ``for'' in dependency parsing.\\
\textbf{Verb}: Complementary relation can be expressed only through verbs without using any preposition. For example, in ``It fits my phone'', subject ``It'' is related to the object ``phone'' via only the transitive verb ``fits''. This pattern has low precision on extraction since almost every sentence has a subject, a verb and an object. We improve the precision of this pattern using the domain knowledge in Section \ref{sec:b}.

\subsection{Dependency Paths for Extraction}

\begin{table*}
\centering
\scalebox{0.95}{
\begin{tabular}{ l | c | c | c }
\hline
Path Type & ID & Path & Example \\ 
\hline
Verb+Prep
& 
1
&
$(\textit{verb, V})\xrightarrow[]{\textit{nmod:cmprel}}(\textit{CETT, N})$
&
It works/V with my phone[\textit{CETT}].
\\
\hline
Noun+Prep
&
2
&
$(\textit{*, N})\xrightarrow[]{\textit{nmod:cmprel}}(\textit{CETT, N})$
&
No problem/N with my phone[\textit{CETT}].
\\
\hline
Adjective+Prep
&
3
&
$(\textit{*, J})\xrightarrow[]{\textit{nmod:cmprel}}(\textit{CETT, N})$
&
It is compatible/J with my phone[\textit{CETT}].
\\
\hline
Determiner+Prep
&
4
&
$(\textit{*, DT})\xrightarrow[]{\textit{nmod:cmprel}}(\textit{CETT, N})$
&
I use this/DT for my phone[\textit{CETT}].
\\
\hline
\multirow{2}{*}{Verb} & 
5
&
$(\textit{verb, V})\xrightarrow[]{\textit{dobj}}(\textit{CETT, N})\xrightarrow[]{\textit{nmod:poss}}(\textit{``my'', PRP\$})$
& It fits my phone[\textit{CETT}]. \\
\cline{2-4} & 
6
&
$(\textit{\textit{``it''/``this''}, DT})\xleftarrow[]{\textit{nsubj}}(\textit{verb, V})\xrightarrow[]{\textit{dobj}}(\textit{CETT, N})$
& It fits iPhone[\textit{CETT}].\\
\hline
\end{tabular}
}
\caption{Summary of dependency paths: \textit{CETT} indicates the complementary entity we want to extract; \textit{verb} indicates any verb for Section \ref{sec:r} or domain-specific verbs for Section \ref{sec:b}.}
\label{table:rule}
\end{table*}

According to the discussed patterns, we implement dependency paths, which are summarized in Table \ref{table:rule}. For patterns with a preposition (e.g., Verb+Prep, Noun+Prep, Adjective+Prep, Determiner+Prep), we use dependency type \textit{nmod:cmprel} to encode all prepositions, because \textit{cmprel} represents \textit{with}, \textit{for}, \textit{in}, \textit{on}, \textit{to}, \textit{inside} and \textit{into} as described in Section \ref{sec:prelim}. Then type \textit{nmod:cmprel} can relate verbs, nouns, adjectives or determiners to the complementary entities. As shown in Example 1 and 2, \textit{nmod:cmprel} can match \textit{nmod:with} and relates the verb ``works'' to the complementary entity ``phone'' for dependency relation 5 in Table \ref{table:dr}. This path is defined as Path 1 in Table \ref{table:rule}.

For pattern Verb, we use dependency type \textit{dobj} to relate a verb to the complementary entity. Since this pattern tends to have low precision, we further constrain the pattern by connecting a \textit{nsubj} relation or a \textit{nmod:poss} relation, as described in Path 5 or Path 6 respectively in Table \ref{table:rule}. For example, ``It fits iPhone'' has the following two dependency relations: \textit{nsubj(``fits'', VBZ, 2, ``It'', PRP, 1)} and \textit{dobj(``fits'', VBZ, 2, ``iPhone'', NNP, 3)}. Path 6 can match these two dependency relations separately and then check the two ``fits''s have the same index \textit{2} in these two dependency relations. So ``iPhone'' tagged as \textit{CETT} can be extracted. 

Finally, these paths may appear multiple times in a sentence. So multiple complementary entities in a sentence can be extracted. For example, ``It works with my phone, laptop and tablet'' has 3 complementary entities. It has the following 3 dependency relations: \textit{nmod:with(``works'', VBZ, 2, ``phone'', NN, 5)}, \textit{nmod:with(``works'', VBZ, 2, ``laptop'', NN, 7)} and \textit{nmod:with(``works'', VBZ, 2, ``tablet'', NN, 9)}. So Path 1 can have 3 matches to extract ``phone'', ``laptop'' and ``tablet''.

Please note that Table \ref{table:rule} does not list all possible dependency paths. For example, complementary entities can also serve as the subject of a sentence: ``My phone likes this card''. We simply demonstrate typical dependency paths and new dependency paths can be easily added into the system to improve the recall.

\subsection{Post-processing}
Since a dependency relation can only handle the relation between two individual words, a complementary entity (labeled by \textit{CETT}) extracted from Subsection B can only contain a single word. In reality, many complementary entities are named entities that represent product names such as ``Samsung/NNP Galaxy/NNP S6/NNP''. Dependency relations usually pick a single noun (e.g., ``S6'') and relate it with other words in the phrase via other dependency relations (e.g., type \textit{compound}). We use the regular expression pattern $\langle \textit{N} \rangle \langle \textit{N}\vert \textit{CD} \rangle \textit{*}$ to chunk a single noun into a noun phrase\footnote{We implement the noun phrase chunker via NLTK: http://www.nltk.org/}. This pattern means one noun (\textit{N}) followed by 0 to many nouns or numbers. Nouns and numbers (model number) are typical POS tags of words in a product name.

\section{Recognition via Domain Knowledge Expansion}
\label{sec:b}
Using the paths defined in Section \ref{sec:r} tends to have low precision (noisy) of extractions since syntactic patterns may not distinguish a complementary relation from other relations. For example, Path 6 can match any sentence with type \textit{dobj}. A sentence like ``It has fast speed'' uses type \textit{dobj} to bring out ``speed'', which is a feature of the target entity itself. To improve the precision, we incorporate category-level domain knowledge (candidate complementary entities and domain-specific verbs) into the extraction process. Those knowledge can help to constrain possible choices of \textit{CETT} and \textit{verb} in dependency paths defined in Section \ref{sec:r}. 

We mine domain knowledge from a large amount of unlabeled reviews under the same category. We get those two types of domain knowledge by bootstrapping them only from general verb \textit{fit} and \textit{work}. We randomly select 6000 reviews for each domain (category) to accumulate enough knowledge (knowledge from reviews of a single target entity may not be sufficient). One important observation is that products under the same domain share similar complementary entities and use similar domain-specific verbs. For example, all \textit{micro SD cards} have \textit{camera}, \textit{camcorder}, \textit{phone}, \textit{tablet}, etc. as their complementary entities and use verbs like \textit{insert} to express complementary relations. But these complementary entities and domain-specific verbs do not make sense for category \textit{tablet stand}. To ensure the quality of the domain knowledge, we utilize several high precision dependency paths. These paths have low recall, so applying them directly to the testing reviews of the target entity has poor performance. High precision paths can leverage big data to improve the precision of other paths in Section \ref{sec:r}.

\begin{table*}
\centering
\scalebox{0.95}{
\begin{tabular}{ l | c | c | c }
\hline
Type & ID & Path & Example \\
\hline
CCE & 
7 &
$(\textit{``fit''/``work'', V})\xrightarrow[]{\textit{nmod:cmprel}}(\textit{CETT, N})\xrightarrow[]{\textit{nmod:poss}}(\textit{``my'', PRP\$})$
& It works with my phone[\textit{CETT}].\\
\hline
\multirow{2}{*}{DSV} & 
8 &
$(\textit{verb, V})\xrightarrow[]{\textit{nmod:cmprel}}(\textit{CETT, N})\xrightarrow[]{\textit{nmod:poss}}(\textit{``my'', PRP\$})$
& I insert[\textit{verb}] the card into my phone[\textit{CETT}]. \\
\cline{2-4} & 
9 &
$(\textit{``this'', DT})\xleftarrow[]{\textit{dobj}}(\textit{verb, V})\xrightarrow[]{\textit{nmod:poss}}(\textit{``my'', PRP\$})$
& This holds[\textit{verb}] my phone[\textit{CETT}] well.\\
\hline 
\end{tabular}
}
\caption{Summary of dependency paths for extracting Candidate Complementary Entities (CCE) and Domain-Specific Verbs (DSV)}
\label{table:bigdatarule}
\end{table*}

\subsection{Exploiting Candidate Complementary Entities}
Knowing category-level candidate complementary entities is important for extracting complementary entities for a target entity under that category. For example, the sentences ``It works in iPhone'', ``It works in practice'' and ``It works in 4G'' have similar dependency relations \textit{nmod:in(``works'', VBZ, 2, ``iPhone''/ ``practice''/ ``4G'', NN, 4)}. But only the first sentence has a mention of a complementary entity; the second sentence has a common phrase ``in practice'' with a preposition ``in''; the third sentence expresses an aspect of the target entity. The key idea is that if we know that \textit{iPhone} is a potential complementary entity under the category of \textit{micro SD card} and ``practice'' and ``4G'' are not, we are confident to extract ``iPhone'' as a complementary entity.

We use Path 7 to extract candidate complementary entities as described in Table \ref{table:bigdatarule}. It has high precision because given a verb like ``fit'' or ``work'', a preposition that relates to another entity and the possessive pronoun ``my'', we are confident that the entity modified by ``my'' is a complementary entity. Lastly, all extracted complementary entities are stored as domain knowledge for each category. 

\subsection{Exploiting Domain-Specific Verbs}
Similarly, knowing category level domain-specific verbs is also important. This is because each category of products may have its own domain verbs to describe a complementary relation. If we only use general verbs (e.g., \textit{fit} and \textit{work}), we may miss many complementary entities that are bring out via domain-specific verbs (e.g., \textit{insert} for \textit{micro SD card} or \textit{hold} for \textit{tablet stand}), and this leads to poor recall rate. In contrast, if we consider all verbs into the paths without distinguishing them as in Section \ref{sec:r}, we may bring in lots of noisy false positives. For example, if the target entity is a \textit{tablet stand}, ``It holds my tablet'' and ``It prevents my finger going numb'' have similar dependency relations ( \textit{dobj(``holds''/``prevents'', VB, 2, ``tablet''/``finger'', NN, 4)} ). The former one has a complementary entity since ``holds'' indicates a functionality that a \textit{tablet stand} can have. The latter one does not have one. So if we know \textit{hold} (we lemmatize the verbs) is a domain-specific verb under the category of \textit{tablet stand} and ``prevents'' is not, we are more confident to get rid of the latter one. Therefore, we design dependency paths to extract high quality domain-specific verbs. This time, candidate complementary entities can help to identify whether a verb has a semantic meaning of \textit{complement}. So we leverage the domain knowledge extracted in Subsection A to extract domain-specific verbs. In the end, we get domain-specific verbs from general seed verbs \textit{fit} and \textit{work}.

Path 8 and 9 in Table \ref{table:bigdatarule} are used to get verbs in pattern Verb+Prep and Verb respectively. These paths also have high precision because given possessive modifier ``my'' modifying a complementary entity or determiner ``this'' indicating a target entity it is almost certain that the verb between them indicates a complementary relation. Then we keep the words tagged by \textit{verb} more than once (to reduce the noise) and store them as domain knowledge. Please note that we do not further expand domain knowledge to avoid reducing the quality of domain knowledge. 

\subsection{Entity Extraction using Domain Knowledge}
We use the same dependency paths in Section \ref{sec:r} to perform extraction. But this time we utilize the knowledge of candidate complementary entities and domain-specific verbs under the same category as the target entity. During matching, we look up candidate complementary entities and domain-specific verbs for tags \textit{CETT} and \textit{verb} respectively. But there is an exception for \textit{CETT}. Since a named entity as a complementary entity may rarely appear again in a large amount of reviews, we ignore such a check if the word covered by \textit{CETT} can be expanded into a noun phrase (more than 1 word) during post-processing. Furthermore, we notice that knowledge about target entities is also useful. For example, ``I insert this card into my phone'' uses ``this'' to bring out the target entities, which may indicate nearby entities are complementary entities. However, knowledge about a target entity may be expanded on reviews of that target entity (test data) rather than reviews under the same category because target entities are not the same under the same category.

\section{Experimental Results}
\label{sec:exp}

\begin{table}
\scalebox{0.95}{
\begin{tabular}{ l || c | c | c | c }
\hline
Product & Revs. & Sents. & Rel. & Revs. w/ Rels.\\
\hline
Stylus & 216 & 892 & 165 & 116 \\
Micro SD Card & 216 & 802 & 193 & 149 \\
Mouse & 216 & 1158 & 221 & 136 \\
Tablet Stand & 218 & 784 & 154 & 115 \\
Keypad & 114 & 618 & 113 & 76 \\
Notebook Sleeve & 109 & 405 & 125 & 84 \\
Compact Flash & 113 & 347 & 99 & 82 \\
\hline 
\end{tabular}
}
\caption{Statistics of the annotated dataset on the number of reviews, the number of sentences, the number of complementary relations and the number of reviews with complementary relations}
\label{table:testingdata}
\end{table}

\begin{table*}
\centering
\scalebox{0.95}{
\begin{tabular}{ l | c c c | c c c | c c c | c c c | c c c }
\hline
\multirow{2}{*}{Product} & 
\multicolumn{3}{ |c| }{NP Chunker} &
\multicolumn{3}{ |c| }{OpenNLP} & 
\multicolumn{3}{ |c| }{UIUC NER} & 
\multicolumn{3}{ |c }{CRF} &
\multicolumn{3}{ |c }{Sceptre} \\
\cline{2-16}&
$\mathcal{P}$&$\mathcal{R}$&$\mathcal{F}_1$&
$\mathcal{P}$&$\mathcal{R}$&$\mathcal{F}_1$&
$\mathcal{P}$&$\mathcal{R}$&$\mathcal{F}_1$&
$\mathcal{P}$&$\mathcal{R}$&$\mathcal{F}_1$&
\multicolumn{3}{ |c }{$\mathcal{P}$@25}\\
\hline
Stylus  &  0.21 & 0.96 & 0.35 & 0.03 & 0.13 & 0.05 & 0.41 & 0.21 & 0.28 & 0.69 & 0.46 & 0.55  &  \multicolumn{3}{ |c }{0.04} \\
Micro SD Card  &  0.26 & 0.99 & 0.41 & 0.04 & 0.14 & 0.07 & 0.34 & 0.39 & 0.36 & 0.85 & 0.47 & 0.6  &  \multicolumn{3}{ |c }{0.16} \\
Mouse  &  0.22 & 0.98 & 0.36 & 0.1 & 0.4 & 0.15 & 0.3 & 0.26 & 0.28 & 0.65 & 0.4 & 0.49  &  \multicolumn{3}{ |c }{0.16} \\
Tablet Stand  &  0.25 & 0.97 & 0.4 & 0.06 & 0.21 & 0.09 & 0.82 & 0.16 & 0.27 & 0.73 & 0.44 & 0.55  &  \multicolumn{3}{ |c }{0.04} \\
Keypad  &  0.2 & 0.98 & 0.33 & 0.05 & 0.21 & 0.08 & 0.4 & 0.25 & 0.31 & 0.63 & 0.24 & 0.35  &  \multicolumn{3}{ |c }{0.04} \\
Notebook Sleeve  &  0.33 & 0.97 & 0.5 & 0.05 & 0.1 & 0.06 & 0.79 & 0.26 & 0.4 & 0.64 & 0.26 & 0.37  &  \multicolumn{3}{ |c }{0.0} \\
Compact Flash  &  0.3 & 0.95 & 0.46 & 0.06 & 0.16 & 0.09 & 0.56 & 0.36 & 0.44 & 0.77 & 0.33 & 0.46  &  \multicolumn{3}{ |c }{0.04} \\
\hline
\multirow{2}{*}{} & 
\multicolumn{3}{ |c| }{``My'' Entity} & 
\multicolumn{3}{ |c| }{\textbf{CER}} & 
\multicolumn{3}{ |c| }{\textbf{CER1K+}} &
\multicolumn{3}{ |c| }{\textbf{CER3K+}} &
\multicolumn{3}{ |c }{\textbf{CER6K+}}\\
\cline{2-16}&
$\mathcal{P}$&$\mathcal{R}$&$\mathcal{F}_1$&
$\mathcal{P}$&$\mathcal{R}$&$\mathcal{F}_1$&
$\mathcal{P}$&$\mathcal{R}$&$\mathcal{F}_1$&
$\mathcal{P}$&$\mathcal{R}$&$\mathcal{F}_1$&
$\mathcal{P}$&$\mathcal{R}$&$\mathcal{F}_1$\\
\hline
Stylus  &  0.5 & 0.54 & 0.52 & 0.35 & 0.89 & 0.5 & 0.89 & 0.64 & 0.75 & 0.88 & 0.69 & 0.77 & 0.86 & 0.71 & \textbf{0.78} \\
Micro SD Card  &  0.63 & 0.51 & 0.56 & 0.39 & 0.8 & 0.52 & 0.81 & 0.64 & 0.71 & 0.79 & 0.66 & 0.72 & 0.8 & 0.67 & \textbf{0.73} \\
Mouse  &  0.54 & 0.37 & 0.44 & 0.35 & 0.91 & 0.5 & 0.69 & 0.69 & 0.69 & 0.66 & 0.7 & 0.68 & 0.66 & 0.72 & \textbf{0.69} \\
Tablet Stand  &  0.58 & 0.43 & 0.49 & 0.41 & 0.84 & 0.55 & 0.68 & 0.39 & 0.5 & 0.75 & 0.69 & 0.72 & 0.75 & 0.72 & \textbf{0.74} \\
Keypad  &  0.54 & 0.46 & 0.5 & 0.33 & 0.92 & 0.49 & 0.66 & 0.67 & 0.66 & 0.67 & 0.73 & 0.7 & 0.69 & 0.82 & \textbf{0.75} \\
Notebook Sleeve  &  0.69 & 0.38 & 0.49 & 0.46 & 0.71 & 0.56 & 0.93 & 0.5 & 0.65 & 0.93 & 0.65 & 0.76 & 0.92 & 0.66 & \textbf{0.77} \\
Compact Flash  &  0.75 & 0.61 & 0.67 & 0.46 & 0.88 & 0.6 & 0.86 & 0.63 & 0.73 & 0.86 & 0.68 & 0.76 & 0.85 & 0.7 & \textbf{0.77} \\
\hline
\end{tabular}
}
\caption{Comparison of different methods in precision, recall and F1-score}
\label{table:comparison}
\end{table*}

\begin{table*}
\centering
\scalebox{0.90}{
\begin{tabular}{ l | c | c | c | c | c }
\hline
Category & 1K(s) & 3K(s) & 6K(s) & Candidate Complementary Entity & Domain-Specific Verbs \\
\hline
Cat:Stylus  &  1.16 &  4.53  &  7.49  & 
ipad 2, tablet, iPhone, Samsung Galaxy 2 &
scratch, match, press, draw, sketch, sign\\
\hline
Cat:Micro SD Card  &  1.23  &  3.67  &  5.58  & 
laptop, psp, galaxy s4, Galaxy tab  & 
add, insert, plug, transfer, store, stick\\
\hline
Cat:Mouse  &  1.61  &  5.1  &  7.71  & 
Macbook pro, laptop bag, MacBook Air &
move, rest, carry, connect, click\\
\hline
Cat:Tablet Stand  &  1.51  &  4.08  &  6.93  & 
Nook, ipad 2, Kindle Fire, Galaxy tab, fire &
rest, insert, stand, support, hold, sit\\
\hline
Cat:Keypad  &  1.25  &  2.93  &  6.17  & 
MacBook, MacBook pro, Mac
&
hook, connect, go, need, use, fit, plug 
\\
\hline
Cat:Notebook Sleeve  &  1.11  &  2.79  &  5.46  & 
backpack, Macbook pro, Lenovo x220 
&
show, scratch, bring, feel, protect
\\
\hline
Cat:Compact Flash  &  1.49  &  3.29  &  6.45  & 
dslr, Canon rebel, Nikon d700
&
load, pop, format, insert, put
\\
\hline

\end{tabular}
}
\caption{Running time (in seconds(s) ) of expanding domain knowledge from 1K, 3K and 6K reviews and samples of candidate complementary entities and domain-specific verbs}
\label{table:bd}
\end{table*}

\subsection{Dataset}
We select reviews of 7 products that have frequent mentions of complementary relations from the Amazon review datasets \cite{McAPanLes15}. We choose accessories because compatibility issues are more frequently discussed in accessory reviews. The products are \textit{stylus}, \textit{micro SD card}, \textit{mouse}, \textit{tablet stand}, \textit{keypad}, \textit{notebook sleeve} and \textit{compact flash}. We select nearly 220 reviews for the first 4 products and 110 reviews for the last 3 products. We select 50\% reviews of the first 4 products as the training data for Conditional Random Field (CRF) (one supervised baseline). The remaining reviews of the first 4 products and all reviews of the last 3 products are test data. We split the training/testing data for 5 times and average the results. We label complementary entities in each sentence. The whole datasets are labeled by 3 annotators independently. The initial agreement is 82\%. Then disagreements are discussed and final agreements are reached. The statistics of the datasets\footnote{The annotated dataset is available on the first author's website \\\url{https://www.cs.uic.edu/~hxu/} } can be found in Table \ref{table:testingdata}. We observe that more than half of the reviews have at least one mention of complementary entities and more than 10\% sentences have at least one mention of complementary entities.

We also utilize the category information in the meta data of each review to group reviews under the same category together. Then we randomly select 1000 (1K), 3000 (3K), 6000 (6K) reviews from each category and use them for extracting domain knowledge. We choose different scales of reviews to see the performance of CER under the help of different sizes of domain reviews and the scalability of the running time of domain knowledge expansion.

\subsection{Compared Methods and Evaluation}
Since the proposed problem is novel, there are not so much existing baselines that can directly solve the problem. Except for CRF, we compare existing trained models or unsupervised methods with the proposed methods.\\
\textbf{NP Chunker}: Since most product names are Noun Phrases (NP), we use the same noun phrase chunker ($\langle \textit{N} \rangle \langle \textit{N}\vert \textit{CD} \rangle \textit{*}$) as the proposed method to extract nouns or noun phrases and take them as names of complementary entity. This baseline is used to illustrate a close to random results.\\
\textbf{OpenNLP NP Chunker}: We utilize the trained noun phrase chunking model from OpenNLP\footnote{https://opennlp.apache.org/} to tag noun phrases. We only consider chunks of words tagged as \textit{NP} as predictions of complementary entities.\\
\textbf{UIUC NER}: We use UIUC Named Entity Tagger \cite{ratinov2009design} to perform Named Entity Recognition (NER) on product reviews. It has 18 labels in total and we consider entities labeled as \textit{PRODUCT} and \textit{ORG} as complementary entities. We use this baseline to demonstrate the performance of a named entity tagger.\\
\textbf{CRF}: We retrain a Conditional Random Field (CRF) model using 50\% reviews of the first 4 products. We use BIO tags. For example, ``Works with my Apple iPhone'' should be trained/predicted as ``Works/O with/O my/O Apple/B iPhone/I''. We use MALLET\footnote{http://mallet.cs.umass.edu/} as the implementation of CRF.\\
\textbf{Sceptre}: We also retrieve the top 25 complements for the same 7 products from Sceptre \cite{McAPanLes15} and adapt their results for a comparison. Direct comparison is impossible since their task is a link prediction problem with different labeled ground truths. We label and compute the precision of the top 25 predictions and assume annotators have the same background knowledge for both datasets. We observe that the predicted products are mostly non-complementary products (e.g., \textit{network cables}, \textit{mother board}) and all 7 products have similar predictions.\\
\textbf{``My'' Entity}: This baseline extracts complementary entities by finding all nouns/noun phrases modified by word ``my'' via dependency type \textit{nmod:poss} (e.g., ``It works with my phone''). The word ``my'' usually indicates a product already purchased, so the modified nouns/noun phrases are highly possible complementary entities. We use path
$$(\textit{CETT, N})\xrightarrow[]{\textit{nmod:poss}}(\textit{``my'', PRP\$})$$
to extract complementary entities and use the same post-process step as CER/CER1K/3K/6K+. \\
\textbf{CER}: This method uses all paths described in Section \ref{sec:r} without using any domain knowledge.\\
\textbf{CER1K+, CER3K+, CER6K+}: These methods incorporate domain knowledge extracted from 1000/3000/6000 domain reviews respectively, as described in both Section \ref{sec:r} and \ref{sec:b}.

We perform our evaluation on each mention of complementary entities and compute precision and recall of extraction. We first count the true positive \textit{tp}, the false positive \textit{fp} and the false negative \textit{fn} of each prediction. For each sentence, one extracted complementary entity that is contained in the annotated complementary entities from the sentence is considered as one count for \textit{tp}; one extracted complementary entity that are not found contributes one count to \textit{fp}; any annotated complementary entity that can not be extracted contributes one count to \textit{fn}. We run the system on an i5 laptop with 4GB memory. The system is implemented using Python. All reviews are preprocessed via dependency parsing \cite{de2008stanford}. 

\subsection{Result Analysis}
Table \ref{table:comparison} demonstrates results of different methods. We can see that CER6K+ performs well on all products. It significantly outperforms CER for each product. This shows that domain knowledge can successfully reduce the noise and improve the precision. More importantly, we notice that using just 3K reviews already gets good performance. This is important for categories with less than 6K reviews. We notice that the F1-scores of CER are close or worse than baselines such as CRF or ``My'' Entity. The major reason of its low precisions is that Path 5 and Path 6 in Table \ref{table:rule} can introduce many false positives as we expected. Please note that removing Path 5 and 6 can increase the F1-score of CER. But to have a fair comparison with CER1K/3K/6K+ and demonstrate the room of improvement, we keep noisy Path 5 and 6 in CER. ``My'' Entity has better precision but lower recall than those of CER baselines since not all complementary entities are modified by ``my''. CRF performs relatively good on these products. But the performance drops for the last 3 products because of the domain adaptation problem. In reality, it is impractical to have training data for each product. Sceptre performs poorly, we guess the reason is that products in ``Items also bought'' are noisy for training labels. The overall recall of UIUC NER is low because many complementary entities (e.g., general entities like \textit{tablet}) are not named entities. Please note that the information of domain knowledge (or unlabeled data) may help other baselines, but all those baselines may not able to adopt domain knowledge easily. The running time of all testing is short (less than 1 seconds), so we omit the discussion here.

Next, we demonstrate the running time of domain knowledge expansion and samples of domain knowledge in Table \ref{table:bd}. We observe that expanding knowledge is pretty fast and scalable as the size of reviews grow. We can see that for each category most entities and verbs are reasonable based on our common sense. For example, for category \textit{Cat:Stylus}, the system successfully detects capacitive screen devices as its candidate complementary entities and most drawing actions as domain-specific verbs.

\subsection{Case Studies}
We notice that category-level domain knowledge is useful for extraction. Knowing candidate complementary entities can successfully remove many words that are not complementary entities or even entities. In the reviews of \textit{micro SD card}, many features such as \textit{speed}, \textit{data}, etc. are mentioned; also, common phrases like ``in practice'', ``in reality'', ``in the long run'' are also mentioned. Handling these cases one-by-one is impractical since identifying different types of false positive examples needs different techniques to identify. But knowing candidate complementary entities can easily remove those false positives. 

Domain-specific verbs such as \textit{draw}, \textit{insert} and \textit{hold} are successfully mined for \textit{stylus}, \textit{micro SD card} and \textit{tablet stand} respectively. Taking \textit{tablet stand} for example, the significant improvement of the precision of CER1K/3K/6K+ comes from taking \textit{hold} as a domain-specific verb. Reviewers are less likely to use general verbs such as \textit{fit} or \textit{work} for \textit{tablet stand}. The reason could be that a \textit{tablet} is loosely attached to a \textit{tablet stand}. So people tend to use ``It holds tablet well'' a lot. However, this sentence has a \textit{dobj} relation that usually relates a verb to an object, which can appear in almost any sentence. Knowing \textit{hold} is a domain-specific verb is important to improve the precision. The major errors come from parsing errors since reviews are informal texts.

\section{Conclusion}
In this paper, we propose the problem of CER. Then we propose an unsupervised method using dependency paths to solve this problem. It further incorporates domain knowledge mined from a large amount of unlabeled reviews to improve its precision. Applications of our work can be found in mining compatible/incompatible products, which is useful for customers, manufacturers and recommender systems. Future directions of our work are mining opinions on complementary relations.

\section*{Acknowledgment}
This work is supported in part by NSF through grants IIS-1526499 and CNS-1626432. We gratefully acknowledge the support of NVIDIA Corporation with the donation of the Titan X GPU used for this research.


%
\bibliographystyle{IEEEtran}
\bibliography{HU_XU}

\end{document}